\documentclass[lettersize,journal]{IEEEtran}
\usepackage{amsmath,amsfonts}
\usepackage{algorithmic}
\usepackage{array}
\usepackage[caption=false,font=normalsize,labelfont=sf,textfont=sf]{subfig}
\usepackage{textcomp}
\usepackage{stfloats}
\usepackage{url}
\usepackage{graphicx}
\usepackage{mathtools}
\usepackage{tikz}
\usetikzlibrary{arrows, positioning, decorations.pathmorphing}
\DeclarePairedDelimiter\abs{\lvert}{\rvert}

\usepackage{amssymb}
\usepackage[font=footnotesize]{caption}
\usepackage{booktabs}
\usepackage{xcolor}
\usepackage{multirow}
\usepackage[english]{babel}

\usepackage[colorlinks=true, allcolors=blue]{hyperref}
\usepackage{amsthm}
\usepackage[noend,ruled,vlined]{algorithm2e}

\def\eg{\emph{e.g., }}
\def\ie{\emph{i.e., }}
\def\etal{\emph{et al.}}

\newcommand{\bbh}[1] {\textcolor{black}{#1}}

\theoremstyle{remark}
\newtheorem*{remark}{Remark}
\begin{document}
\title{DeepGraphONet: A Deep Graph Operator Network to Learn and Zero-shot Transfer the Dynamic Response of Networked Systems}
\author{%
	\IEEEauthorblockN{%
		Yixuan Sun,
		Christian Moya,
		Guang Lin,
		Meng Yue%
	}\\%

	\vspace{1em}
	\thanks{Y. Sun and G. Lin are with the School of Mechanical Engineering, Purdue University, West Lafayette, IN. E-mail:~\{\texttt{yixuan-sun, guanglin}\}\texttt{@purdue.edu}.}
	\thanks{C. Moya and G. Lin are with the Department of Mathematics, Purdue University, West Lafayette, IN. e-mail:~\{\texttt{cmoyacal, guanglin}\}\texttt{@purdue.edu}.}%
	\thanks{M. Yue is with the Interdisciplinary Science Department, Brookhaven National Laboratory, Upton, NY. e-mail:~\texttt{yuemeng@bnl.gov}.}%
}

%


\maketitle

\begin{abstract}
This paper develops a Deep Graph Operator Network~(DeepGraphONet) framework that learns to approximate the dynamics of a complex system (\eg the power grid or traffic) with an underlying sub-graph structure. We build our DeepGraphONet by fusing the ability of (i) Graph Neural Networks~(GNN) to exploit spatially correlated graph information and (ii) Deep Operator Networks~(DeepONet) to approximate the solution operator of dynamical systems. The resulting DeepGraphONet can then predict the dynamics within a given short/medium-term time horizon by observing a finite history of the graph state information. Furthermore, we design our DeepGraphONet to be resolution-independent. That is, we do not require the finite history to be collected at the exact/same resolution. In addition, to disseminate the results from a trained DeepGraphONet, we design a zero-shot learning strategy that enables using it on a different sub-graph. Finally, empirical results on the (i) transient stability prediction problem of power grids and (ii) traffic flow forecasting problem of a vehicular system illustrate the effectiveness of the proposed DeepGraphONet.
\end{abstract}


\renewcommand{\figurename}{Fig.}

\section{Introduction} \label{intro}
\IEEEPARstart{N}{etworked} dynamical systems are ubiquitous in science and engineering, \eg the power grid or traffic networks. To simulate and predict the dynamics of such systems, researchers have developed sophisticated, high-fidelity numerical schemes that can accurately solve the corresponding governing equations. However, for tasks requiring multiple forward simulations, \eg optimization, uncertainty quantification, and control, these high-fidelity schemes may become prohibitively expensive~\cite{iserles2009first}.

Deep learning techniques can potentially address this computational cost problem by acting as a fast surrogate model trained using data. As a result, many deep learning techniques have been proposed in the literature to simulate and predict complex dynamical systems. For example, one can use traditional neural networks, such as fully connected neural networks or recurrent neural networks~(RNN), for time-dependent prediction tasks~\cite{li2020machine, wilms2019exploiting, sun2021data}. Such networks have the goal of learning  (i) the underlying governing equations~\cite{brunton2016discovering,kaiser2018sparse, schaeffer2017learning} or (ii) the future (\ie the next state) dynamic response~\cite{qin2019data, raissi2018multistep}. However, such traditional neural networks often require a massive amount of data (with a fixed resolution) to learn the system's dynamic response to a particular operating condition. For other operating conditions (\eg different initial conditions), one often needs to retrain the model with a different dataset. Such a limitation has prevented the wider use of traditional neural networks for solving complex dynamical systems in science and engineering, where the data is scarce and expensive to collect.

To address such a limitation, several works have proposed to learn the solution operator (\ie a mapping between infinite-dimensional spaces) of complex dynamical systems using, for example, Deep Operator Networks~(DeepONet)~\cite{lu2021learning}, Graph Neural Operators~(GNO)~\cite{li2020neural}, or Fourier Neural Operators~(FNO)~\cite{li2020fourier}. In particular, the DeepONet, introduced in the seminal paper~\cite{lu2021learning} and developed based on the universal approximation theorem for nonlinear operators~\cite{tianping_chen_universal_1995}, has demonstrated remarkable accuracy and generalization capability for learning the solution operator of non-autonomous systems.

We used the DeepONet framework in a recent study~\cite{moya2022deeponet} to learn the dynamic response of the power grid after a disturbance without using recurrence and fixed resolution, requirements typically adopted by RNNs. However, our work only considered the dynamic response at the bus level. Thus, it neglects the possibly rich information from the interaction between a bus and its neighbors. Such a spatial correlation is critical for learning the solution operator of networked dynamical systems, \eg the power grid or traffic networks.

To enable learning the solution operator of complex networked dynamical systems, we propose, in this paper, the Deep Graph Operator Network~(DeepGraphONet) framework. DeepGraphONet is a muti-input multi-output Deep Operator Network that learns the dynamic response of complex networked systems by exploiting spatially correlated information from the given underlying graph or sub-graph structure.

\subsection{Related Work} \label{rel}
\textit{Learning dynamical systems.} Many works have proposed using machine and deep learning to learn unknown dynamical systems from time-series data. In particular, we classify such works into learning the system's (i) governing equations~\cite{brunton2016discovering, schaeffer2017learning, zhang2018robust} and (ii) future response~\cite{qin2019data, raissi2018multistep}. For example, in~\cite{brunton2016discovering}, the authors use a dictionary of functions to learn a sparse representation of the system's governing equations. Zhang \etal~\cite{zhang2018robust} adopted Bayesian sparse regression to identify differential equation terms from a large pool of candidates with error bars.

On the other hand, Qin~\etal~\cite{qin2019data} trains a neural network to learn the next state response of the system given the current state. The framework then can predict the system's future response by recursively using the trained network. In~\cite{raissi2018multistep}, the authors proposed a multistep method with a feed-forward neural network to approximate the dynamical system response. Most of the above works can effectively learn the dynamical system for a single operating condition. However, the above works will require a prohibitive amount of data and training resources to learn the system's response to many operating conditions. To alleviate such limitations, we will use the novel framework of deep operator learning in this work.

\textit{Deep operator learning.} Traditional deep learning techniques~\cite{goodfellow2016deep} focus on approximating the mapping between Euclidean spaces. However, these traditional techniques may not be adequate for learning the solution operator of a complex dynamical system. To learn such an operator, several works have proposed designing operator learning frameworks based on deep neural networks~\cite{lu2021learning, li2020neural, li2020fourier, winovich2019convpde}.

We build our work on the Deep Operator Network~(DeepONet) framework introduced in~\cite{lu2021learning}. DeepONet is a neural network architecture that can learn nonlinear operators by using (i) a Branch network to process input information and (ii) a Trunk network to process query locations within the output function domain. Then, one computes the output function at a query location by merging the features from both the Branch and Trunk nets using a dot product. DeepONet has demonstrated its exceptional approximation and generalization capabilities using a low amount of data for problems in power engineering~ \cite{moya2022deeponet}, electroconvection, and multi-physics tasks~\cite{cai2021deepm}, or material science tasks~\cite{goswami2021physics}. DeepONet, however, is only a single-input single-output operator framework. Thus, one cannot directly use DeepONet for networked dynamical systems.

In~\cite{jin2022mionet}, the authors proposed MIONet, a multi-input function single-output function DeepONet to alleviate the multiple-input problem. Nevertheless, the MIONet framework fails to consider the spatial correlation among input functions of a networked system and only produces a single output function. Our work effectively alleviates both limitations by incorporating a Graph Neural Network within the DeepONet framework.

\textit{Graph Neural Networks.} In traditional deep learning, the training data, \eg time series, tabular-like data, and images, is well-structured, and one assumes it belongs to the Euclidean space. However, we cannot naturally assume that other forms of data, such as power grid and traffic data, belong to the Euclidean space. As a result, researchers developed Graph Neural Networks (GNN)~\cite{wu2020comprehensive} to capture the spatial correlation of data with an underlying graph representation.

At the early stage, GNNs were used within recursive schemes~\cite{572108, inproceedings} to reach steady node states for subsequent tasks. These schemes, however, suffered from high-computational costs and were limited to some specific graph structures. To alleviate such issues, other works~\cite{estrach2014spectral,hammond2011wavelets, kipf2016semi} build on graph signal processing to develop graph spectral convolution networks and their localized variants. In graph spectral convolution, one performs the graph convolution operation in the Fourier domain. Such spectral convolution requires the eigendecomposition of the graph Laplacian. Spectral convolution, however, is expensive and does not allow transferring to another graph with a different topology~\cite{Wu2021, Zhou2020}. This limitation prevents the trained GNNs from achieving zero-shot learning.

On the other hand, similar to standard convolutional neural networks, several works~\cite{atwood2016diffusion,niepert2016learning} proposed non-spectral and spatial convolution directly defined on the graph. In particular, the spatial convolution locally operates on nodes, and, thus, one can use it on different graphs. Furthermore, by stacking spatial convolutional layers, we can effectively process information from nodes located more than one hop away.

To learn with graph data efficiently and effectively, the authors in~\cite{gilmer2017neural, velivckovic2017graph, hamilton2017inductive} employed the spatial graph convolution within a message passing framework. The main idea of such a framework is to update the node state information by aggregating information from neighboring nodes, followed by a neural network-based transformation. In this work, we adopt such a message passing framework and, in particular, the framework from GraphSage~\cite{hamilton2017inductive}, to process the input information to the Deep Operator Network.

This work aims at learning the dynamic evolution of graph signals using a fixed graph structure. Many studies~\cite{guo2019attention,Li2016, li2017diffusion, zhao2019t,zhu2021ast} have combined deep learning techniques for time series (\eg RNNs~\cite{li2017diffusion, Li2016} and CNNs~\cite{guo2019attention, zhao2019t, zhu2021ast}) with GNNs to produce time-dependent predictions that can take into account spatial correlations. However, such methods may (i) lack the flexibility and (ii) require high computational-cost to learn the solution operator of a networked dynamical system. Thus, in our work, we will employ the deep operator learning framework to learn the parametric mapping between graph functions directly.

\subsection{Our Work} \label{ssec:our-work}

The objectives of this work are two-fold.
\begin{enumerate}
    \item \textit{Approximating the solution operator.} We aim to derive a deep learning-based method to approximate the solution operator of a system in which the dynamics evolve within an underlying sub-graph structure. For instance, we aim to use our method to learn the dynamics of a control area within a large-scale power grid or the traffic dynamics of a city's neighborhood.
    \item \textit{Zero-shot learning}. We aim to apply the trained proposed model directly to approximate the solution operator on unseen graphs or sub-graphs with different structures. In other words, we aim to achieve zero-shot learning. For instance, we aim to train the model on a small sub-graph and then use it to make predictions on a larger graph with high accuracy.
\end{enumerate}

We detail the contributions of this work next.
\begin{itemize}
    \item We first build (in Section~\ref{method}) a Deep Graph Operator Network~(DeepGraphONet) framework that learns to approximate the solution operator of a dynamical system with an underlying connected sub-graph structure. DeepGraphONet fuses the ability of (i) Graph Neural Networks~(GNN) to exploit the graph information and (ii) DeepONets~\cite{lu2021learning} to approximate the solution operator of nonlinear systems. The proposed DeepGraphONet takes as inputs (i) a finite history of the graph state information and (ii) the desired query location with an arbitrary resolution for short/medium term prediction. Compared to the vanilla DeepONet~\cite{lu2021learning}, the proposed DeepGraphONet is (i) a multi-input multi-output framework and (ii) resolution-independent in the input function; that is, we do not require the Branch sensors to be fixed.
    \item We then propose (in Section~\ref{exp}) a \textit{zero-shot learning} strategy that exploits the property of message passing GNNs to enable directly using a trained DeepGraphONet on a different graph or sub-graph for the same task.
    \item Finally, we verify the efficacy of the DeepGraphONet on (i) the transient stability prediction problem of the IEEE 16-machine 68-bus system and (ii) the traffic dynamics forecasting problem using the METR-LA dataset, which contains traffic information measured using loop detectors in the highway of Los Angeles County~\cite{jagadish2014big}.
\end{itemize}

We organize the rest of this paper as follows. Section~\ref{sec:problem_formulation} introduces the problem of approximating the solution operator of a dynamical system with an underlying sub-graph structure. The DeepGraphONet framework to approximate such a solution operator is detailed in Section~\ref{sec:model}. We present the resolution-independent DeepGraphONet and our zero-shot learning strategy in Section~\ref{sec:res-indepdent} and Section~\ref{sec:zero-shot}, respectively. Numerical experiments are used to illustrate the efficacy of DeepGraphONet in Section~\ref{exp}. Finally, Section~\ref{sec:dis} discusses our future work and Section~\ref{sec:conclusion} concludes the paper.





\section{Problem Settings} \label{sec:problem_formulation}




We consider a complex networked dynamical system. We model its (i) \textit{networked structure} using the undirected graph~$G=(V, E)$, where $V$ is the set of $|V|$ nodes and $E$ is the set of $|E|$ edges, and (ii) \textit{dynamics} using the initial value problem (IVP):
\begin{align}
\begin{aligned}
    \frac{d}{dt} \mathbf{x}(t) &= f(\mathbf{x}(t); G) \\
    \mathbf{x}(t_0) &= \mathbf{x}_0 \label{eqn: dynamics}
\end{aligned}
\end{align}
Here $\mathbf{x}(t) \in \mathcal{X} \subset \mathbb{R}^{d \times|V|}$ is the graph-valued state function, $\mathbf{x}_0 \in \mathcal{X}$ the initial condition, and $f: \mathcal{X} \to \mathcal{X}$ the \textit{unknown} vector field. For simplicity, we assume each node's state has the same dimension~$d=1$ and, with some abuse of notation, we make explicit the dependency of the dynamics~$f$ on the underlying graph structure~$G$. We also let for future work the case when we know an approximate model of the dynamics~$f_\text{approx} \approx f$. 

\textit{Solution operator}. Our goal is to approximate the dynamic response of~\eqref{eqn: dynamics} described via the solution operator (also known as flow map): 
\begin{align*}
    \mathcal{F}(\mathbf{x}(t_0); G)(t) &\equiv \mathbf{x}(t) = \mathbf{x}(t_0) + \int^t_{t_0} f(\mathbf{x}(\beta); G)d\beta,
\end{align*}
for all $t \in [t_0,t_f]$, where $[t_0,t_f] \subset [t_0,\infty)$ denotes a finite time interval where the solution operator exists and is unique.

For large-scale networked dynamical systems, \ie when $|V| \gg 1$, learning the above solution operator may be prohibitively expensive. Moreover, one may not have the access to all graph states and may be interested in learning the dynamics of a sub-region within the network, \ie a \textit{sub-graph}. For example, for power grids, we may want to learn the dynamic response of a Control Area, or, for traffic networks, we may want to approximate the dynamic response of a city or a district.

To this end, let us split the graph-valued state function as $\mathbf{x} = \left(\mathbf{x}_S, \mathbf{x}_{S^c}\right)$. Here, $S=(V_S, E_S)$ denotes the sub-graph region of interest within the network, where $V_S \subset V$ is a \textit{connected set} of nodes within $V$ and $E_s \subset E$, and $S^c$ denotes the complement of~$S$. To describe the dynamics of $\mathbf{x}_S(t)$, we employ the Mori-Zwanzig formalism~\cite{chorin2009stochastic, chorin2002optimal}, which yields
\begin{align*}
    \frac{d}{dt}\mathbf{x}_S(t) = R(\mathbf{x}_S(t);S) + \int_{t_0}^t \mathcal{K}(\mathbf{x}_S(t - \beta), \beta)d\beta + \mathcal{N}(\mathbf{x}_0)   
\end{align*}
In the above, the first term~$R$ is the Markovian term, which depends on the current value of~$\mathbf{x}_S$; the second term is the \textit{memory} integral, which depends on the values of the state and input variables from the initial time $\beta=t_0$ to the current time~$\beta=t$. This memory integral involves~$\mathcal{K}$, which is commonly known as the memory kernel. Finally, the third term is the ``noise'' term, which depends on the initial condition~$\mathbf{x}_0$.

\textit{Decaying memory assumption.} Learning the memory integral from the initial time~$\beta = t_0$ is a challenging task. To alleviate such a challenge, we assume the memory decays over time, that is, there exists~$t_M \in (t_0, t)$ (we treat~$t_M$ as a \textit{hyperparameter} in this paper) such that we can neglect the effect of the memory for all $\beta \in [t_0, t-t_M)$. Formally, this corresponds to the following approximation of the memory integral. For any given~$\epsilon>0$, there exists~$t_0 < t_M < t$ such that:
\begin{align}
    \abs*{\int_{t_0}^t \mathcal{K}(\mathbf{x}_S(t - \beta), \beta)d\beta - \int_{t_0}^{t_M}\mathcal{K}(\mathbf{x}_S(t - \beta), \beta)d\beta} < \epsilon \label{ineqn:1}
\end{align}

\textit{Discrete representation of the integral.} In practice, one computes the truncated integral using some discretized scheme (\eg quadrature) such that for any given~$\kappa>0$, we have
    \begin{align}
        \begin{split}
	     &\left| \int_{t_0}^{t_M} \mathcal{K}(\mathbf{x}_S(t - \beta), \beta)d\beta \right. \\  &~~~~~\qquad~\qquad\left. -Q(\mathbf{x}_S(t - \tau_1), \dots, \mathbf{x}_S(t - \tau_m)) \vphantom{\int_{t_0}^{t_M}} \right| < \kappa,  
	    \end{split}
	    \label{ineqn:2}
    \end{align}
where the memory partition of size~$m$ (we treat this memory resolution~$m$ as a \textit{hyperparameter} in this paper) is \textit{arbitrary} and satisfies $0 \le \tau_1, \ldots, \tau_m \le t_M$ and $\{t - \tau_i\}_{i=1}^m \subset [t-t_M, t]$.

These decaying memory and discrete representation assumptions lead to the following approximate dynamics of~$\mathbf{x}_S$:
\begin{align} \label{eq:approx_dyn}
    \frac{d}{dt}\mathbf{x}_S(t) = R(\mathbf{x}_S(t)) + Q(\mathbf{x}_S(t - \tau_1), \dots, \mathbf{x}_S(t - \tau_m)). 
\end{align}
In this paper, we employ the \textit{local solution operator} within the arbitrary interval $[t, t + h_n]$, $h_n \in [0, h] $, where $h$ is the predicting horizon, and we treat it as a hyperparameter in this paper. For all $t > t_0$, the local memory sensor locations satisfy $\{t - \tau_i\}^m_{i=1} \subset [t - t_M, t]$. With this local memory, the local solution operator is:
\begin{align*}
	\begin{split}
    \tilde{\mathcal{F}}(\mathbf{x}_S(t), & \{\mathbf{x}_S(t - \tau_i)\}^{m}_{i=1}; S)(h_n) =  \mathbf{x}_S(t)\\ 
    &+ \int_{t}^{t + h_n} R(\mathbf{x}_S(\beta); S)d\beta\\
	&~~~~~+ Q(\mathbf{x}_S(t - \tau_1), \dots, \mathbf{x}_S(t - \tau_m)). 
	\end{split}
\end{align*}
 In the above, we emphasized that the solution operator depends on the finite history of the sub-graph state information~$\mathbf{x}_S(t)$ within the time interval $[t-t_M, t]$.

Our goal in this paper is to design a Deep Graph Operator Network~(DeepGraphONet)~$\mathcal{F}_\theta$, with trainable parameters $\theta \in \mathbb{R}^p$, to approximate the solution operator $\tilde{\mathcal{F}}$ for all $h_n \in [0,h]$. To control the approximation power of the proposed DeepGraphONet, we will use \textit{two} hyperparameters: the memory's (i) size~$t_M$ and (ii) predicting horizon~$h$.

\section{Methods} \label{method}
This section describes the proposed \textit{Deep Graph Operator Network}~(DeepGraphONet)~$\mathcal{F}_\theta$ to approximate the solution operator~$\tilde{\mathcal{F}}$.
\begin{figure*}[t]
    \centering
    \includegraphics[width=.9\linewidth]{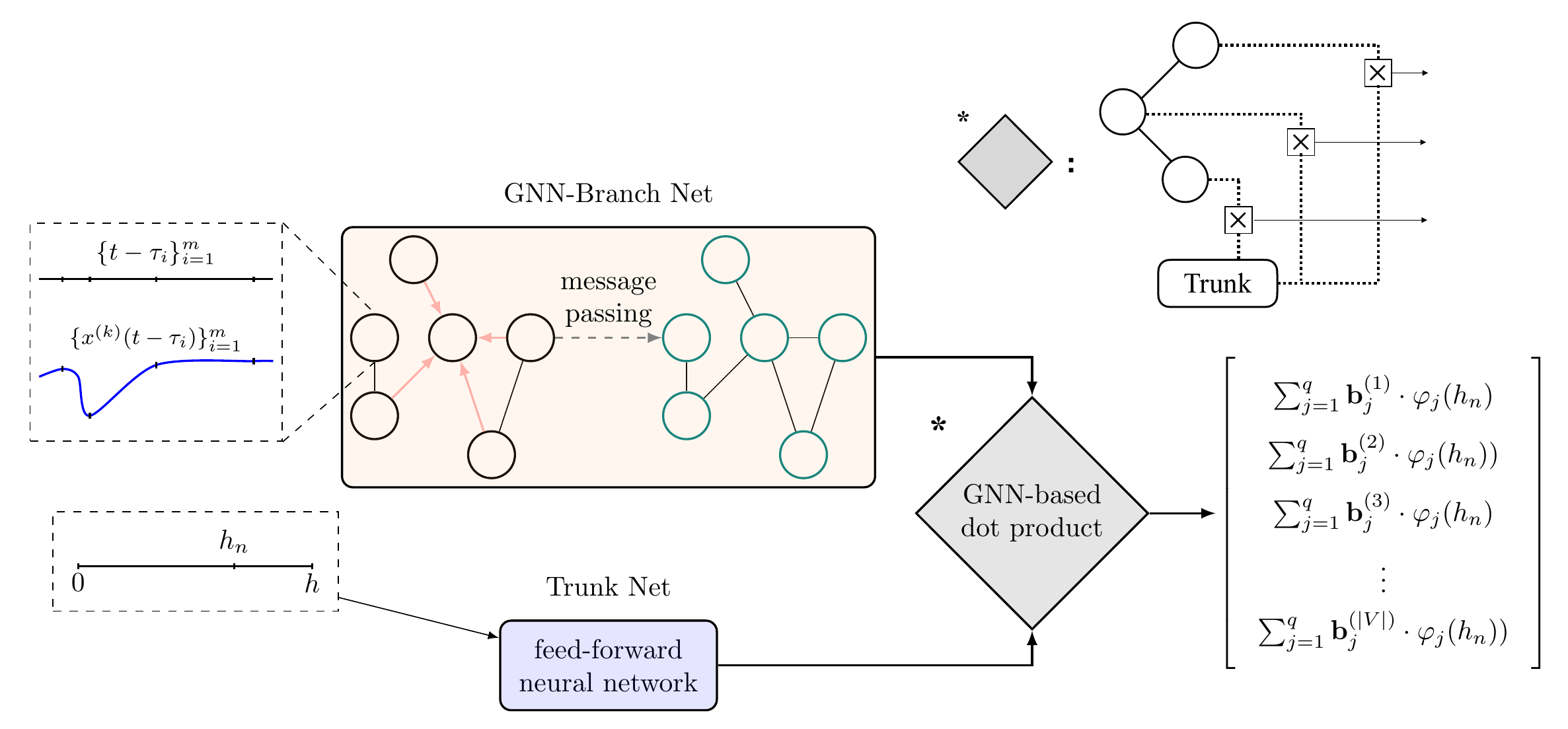}
    \caption{The DeepGraphONet architecture. The \textit{Branch Net} (i) is a Graph Neural Network that uses message passing to learn the nodes' representation on the networked system's graph~$S$ and (ii) takes as input the graph state signals~$\{\mathbf{x}_S(t - \tau_i)\}_{i=1}^m$, $\tau \in [0, t_M]$. The \textit{Trunk Net} is a feed-forward neural network that takes as input the query location~$h_n$ within the prediction horizon~$[0, h]$. We obtain the DeepGraphONet's output~$\mathcal{F}_{\theta}(\{\mathbf{x}_S(t - \tau_i)\}_{i=1}^m; S)(h_n)$ by applying the GNN-based dot product between the GNN-Branch Net and the Trunk Net. The GNN-based dot product performs the standard dot product between each node in the Branch Net-produced graph and the Trunk Net.}
    \label{network}
\end{figure*}
\subsection{The Deep Graph Operator Network} \label{sec:model}
Building on the DeepONet, introduced in~\cite{lu2021learning}, we propose DeepGraphONet~$\mathcal{F}_\theta$, a multi-input multi-output Deep Graph Operator Network, which consists of two neural networks (see Fig.~\ref{network}): the \textit{Branch} network and the \textit{Trunk} network. 

\textit{The Branch network.} The \textit{Branch} net is a \textit{Graph Neural Network}~(GNN) that processes the finite history of the graph-state information. 
The \textit{Branch} maps the graph-state information
$\{\mathbf{x}_S(t - \tau_i)\}_{i=1}^m$, where $\mathbf{x}_S(t -\tau_i) \in \mathbb{R}^{|S|}$, to the graph-based coefficients~$\mathbf{b} \in \mathbb{R}^{q \times|S|}$.

To build the proposed GNN-based Branch network, we use a collection of \textit{message passing convolutional graph layers}, which we adopted from~\cite{hamilton2017inductive}. For all $i \in S$, the forward pass of the corresponding message passing layer reads:
\begin{align*}
	\begin{split}
    x_i'(t) = \sigma(W_1 x_i(t) + W_2 \cdot \text{mean}_{j \in \mathcal{N}_i}(x_j(t))),
    \end{split}
\end{align*}
where $x_i(t)$ is the $i$th node's state at time $t$, $\mathcal{N}_i$ is the set of nodes adjacent to the $i$th node, $W_1$, $W_2$ are trainable weights, and $\sigma$ is a non-linear activation function.

By stacking these message passing layers, we can transport information from nodes located \textit{two or more} hops away to the $i$th node. Such a process allows the network to learn automatically \textit{spatial dependencies} among nodes. Furthermore, since our goal is to obtain the dynamic response of all nodes,  we do not use a readout (pooling) in the proposed GNN-based Branch net.

\begin{remark}
\textit{Fixed v.s. resolution-independent sensors.} If the sensor locations used to discretize the input representing the finite history of graph-state information $\{\mathbf{x}_S(t - \tau_i)\}_{i=1}^m$ are fixed (as in~\cite{lu2021learning}), then we denote the proposed framework as the \textit{standard} DeepGraphONet. If, on the other hand, the sensor locations can change over time, then we denote the proposed framework as the \textit{resolution-independent} DeepGraphONet (see Section~\ref{sec:res-indepdent} for more details).
\end{remark}

\textit{The Trunk network.} The \textit{Trunk} net is a multi-layer perceptron~(MLP) that maps a given query location~$h_n$, within the prediction horizon~$[0,h]$, to a collection of \textit{trainable} Trunk basis functions:
$$\varphi:=(\varphi_1(h_n),\varphi_2(h_n), \ldots,\varphi_q(h_n))^\top \in \mathbb{R}^q.$$

Finally, we compute the DeepGraphONet's output for each node $i \in S$ by merging the corresponding Branch coefficients~$\mathbf{b}^{(i)}$ with the Trunk basis functions~$\varphi$ using the dot product:
$$\mathcal{F}_{\theta}^{(i)}(\{\mathbf{x}_S(t - \tau_i)\}_{i=1}^m;S)(h_n) = \sum_{j=1}^q \mathbf{b}_{j}^{(i)} \cdot \varphi_j(h_n).$$

\textit{Training the DeepGraphONet $\mathcal{F}_{\theta}$}. To train the parameters~$\theta$ of the proposed DeepGraphONet, we minimize, using gradient-based optimization schemes (\eg Adam~\cite{kingma2014adam}), the loss function:
\[\mathcal{L}(\theta; \mathcal{D}) = \frac{1}{N} \sum_{k=1}^N \abs*{ \mathbf{x}_S^k(t + h_n^k) - \mathcal{F}_{\theta}(\{\mathbf{x}_S^k(t - \tau_i)\}_{i=1}^m)(h_n^k)}_1.\]
using the dataset~$\mathcal{D}$ of $N:=|\mathcal{D}|$ triplets:
$$\mathcal{D}:=\left\{\{\mathbf{x}_S^k(t - \tau_i)\}_{i=1}^m, h_n^k, \mathbf{x}_S^k(t + h_n^k)\right\}_{k=1}^N.$$
generated using the true solution operator~$\mathcal{F}$.

\subsection{Resolution-independent DeepGraphONet} \label{sec:res-indepdent}
\begin{figure}[t]
    \centering
    \includegraphics[width=\linewidth]{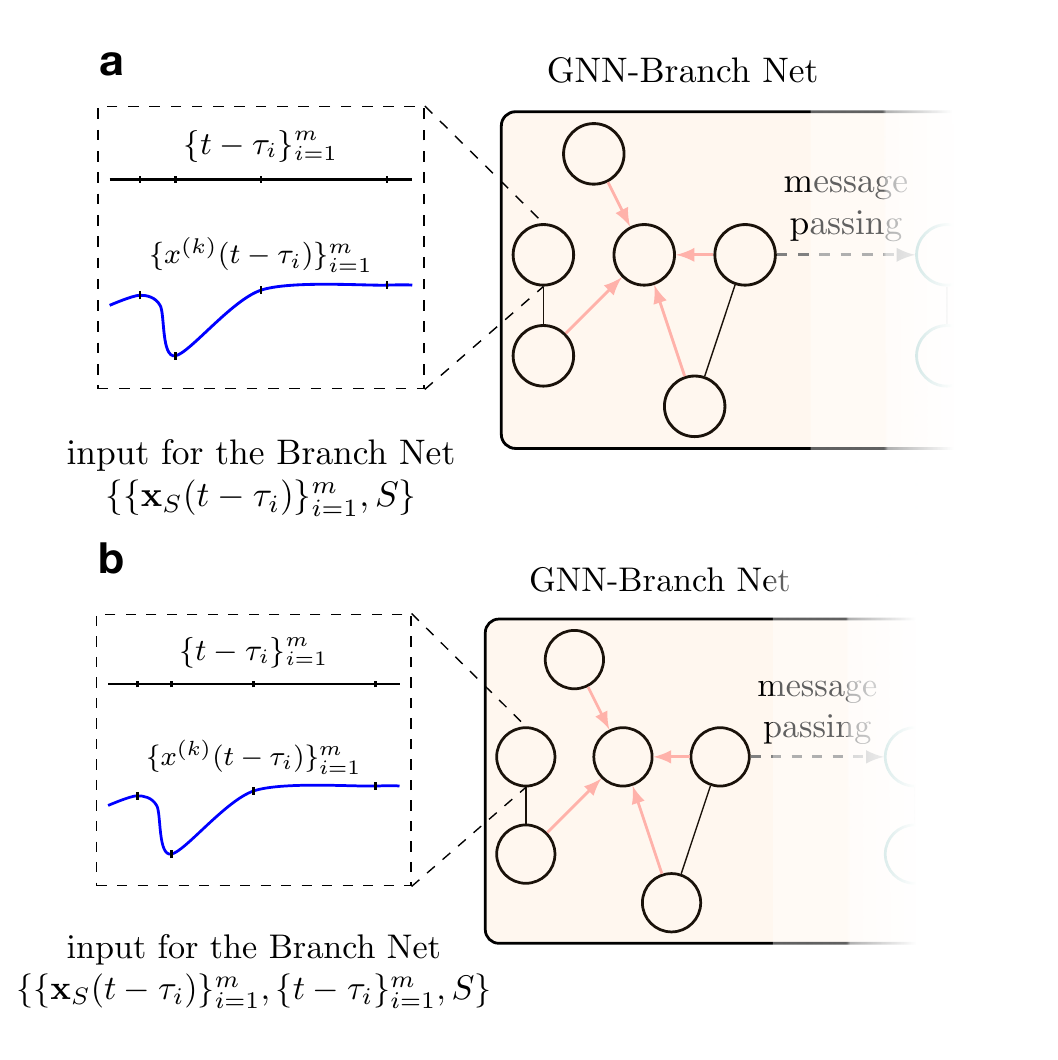}
    \caption{The input for the Branch Net in standard and resolution-independent DeepGraphONet. \textbf{a.} shows the input for the Branch Net in the \textit{standard} DeepGraphONet, where the network takes the input function values at a fixed set of sensors. \textbf{b.} presents, in a \textit{resolution-independent} DeepGraphONet, the Branch Net takes both input function values and their locations. The fixed set of sensors requirement is eliminated.}
    \label{fig:resolutionIndependent}
\end{figure}
The vanilla DeepONet, introduced in~\cite{lu2021learning}, effectively and accurately approximates single-input single-output solution operators~$\mathcal{F}$. Thus, DeepONet can predict at arbitrary locations within the output function domain~$[0, h]$. However, the main limitation of the vanilla DeepONet is that it requires a fixed set of sensors within the input function domain~$[t-t_M, t]$. This requirement forces the network to take new data points with the same resolution, which, in turn, prevents the flexible application of the trained network, \eg in scenarios where (i) sampling may be inexact or (ii) recursive prediction is needed.

To address such a limitation, we propose a simple, intuitive solution that allows using resolution-independent sensors, as shown in Fig. \ref{fig:resolutionIndependent}. More specifically, we enable the GNN-based Branch network to take as inputs not only the graph-state memory input $\{\mathbf{x}_S(t - \tau_i)\}_{i=1}^m$, but also the corresponding, possibly time-variant, sensor locations. To this end, for each node~$i \in S$, we construct the input function by concatenating the node-state function values~$\{\mathbf{x}_S(t - \tau_i)\}_{i=1}^m$ with its corresponding sensor location~$\{t - \tau_i\}_{i=1}^m$. These function-location value pairs provide complete information about the input, enabling us to relax the fixed sensors' requirement. In section \ref{exp}, we will demonstrate that the \textit{resolution-independent} DeepGraphONet does not cause performance degradation when compared to the \textit{standard} DeepGraphONet.
 
\subsection{Zero-Shot Learning Scheme} \label{sec:zero-shot}
In this paper, we build the proposed DeepGraphONet framework to be versatile. That is, once we have trained the DeepGraphONet~$\mathcal{F}_{\theta^*}$ using the graph~$S$, we expect to use it with high accuracy on a different graph~$S' \neq S$ (with possibly different number of nodes) for the same task.

Such a \textit{zero-shot learning} capability of the proposed DeepGraphONet is achieved immediately due to the message-passing nature of the GNN-based Branch network. In particular, we can directly apply DeepGraphONet to $S'$ because the trainable weights of the GNN-based Branch depend on the node feature dimension and not on the adjacency matrix. Thus, the proposed DeepGraphONet is inductive, and we may use it beyond the graph~$S$ used during training. In Section~\ref{exp}, we will show that DeepGraphONet can effectively approximate the solution operator of networked dynamical systems within sub-graph structures~$S'$ not used during training; thus, achieving zero-shot learning.

\section{Experiments} \label{exp}

To evaluate the accuracy and effectiveness of the proposed DeepGraphONet, we test it on \textit{two} different tasks: (i) predicting the transient response of power grids~(Section \ref{sec:power_exp}) and (ii) forecasting traffic on highway networks~(Section \ref{sec:traffic_exp}). In the experiments, we use $t_M$, $m$, $h$, and $h_n$ to denote the memory length, number of sensors within the memory, prediction horizon, and query location within the prediction horizon, respectively. We treat $t_M$ and $h$ as hyperparameters in this paper and explore the model performance.

\textit{Neural networks and implementation protocols.} We use the same neural network architecture to build two DeepGraphONets: the (i) \textit{standard} DeepGraphONet, which has fixed sensors as in~\cite{lu2021learning}, and (ii) \textit{resolution-independent DeepGraphONet}, which we proposed in this paper (see Fig. \ref{fig:resolutionIndependent}) to alleviate these fixed sensors constraint. To build the DeepGraphONets' architecture, we proceed as follows. The \textit{Branch Net} uses twenty \textit{message passing} graph convolution layers, and the \textit{Trunk Net} uses five feed-forward layers. The output dimension is 100 neurons for both the Branch and Trunk Nets.

We prepared the data to align with the DeepGraphONet's formulation, where we set a memory partition and an output function domain bound $h$. With regard to the resolution-independent DeepGraphONet, with the memory partition size $m$, we randomly sampled $\lfloor \frac{m}{2} \rfloor$ sensors and stacked them with their locations as additional feature channels for the \textit{Branch Net}. 

Algorithm \ref{alg:inference} shows the procedure of using trained DeepGraphONet to predict the complete future trajectories. With the trained network, we use the initial memory and partition to produce predictions covering the entire prediction horizon. We then repeat this process with the newly observed \textit{ground truth} as the memory for the next prediction horizon until reaching the end of the trajectory.

For example, in the power grid problem, we have a trained DeepGraphONet with memory length $t_M=200~ms$ and prediction horizon $h=20~ms$. In testing time, we want to produce predictions of the entire trajectory of size $700~ms$ after the initial memory. We start with using the initial memory and the trained network to make predictions for the next 20~$ms$ and save the predictions to a list. We then use the \textit{new observations} to produce predictions for another 20~$ms$ until reaching the end of the trajectory. Finally, we concatenate the saved predictions as the predicted complete trajectory.

We implemented the DeepGraphONet in \texttt{PyTorch} and trained and tested it with an Nvidia GPU. The source code of the model can be accessed on \href{https://github.com/cmoyacal/graph-deepOnet}{GitHub} \footnote{The repository will be public upon the acceptance of the manuscript}. We measured and compared the performance of all trained models in terms of $L_1$ relative error (\%), which can be interpreted as the percentage absolute deviation from the true values.

\begin{algorithm}[t]
\DontPrintSemicolon
\SetAlgoLined
\textbf{Require:} 
system graph $S$;
trained DeepGraphONet~$\mathcal{F}_{\theta^*}$;
hyperparameters~($t_M$, $m$, $h$, $h_n$);
initial local memory $\{\mathbf{x}_S(t - \tau_i)\}_{i=1}^m$;
complete trajectory length $L$; 

initialize an empty list for saving the predictions $C$;

let $N = \lceil\frac{(L - t_M)}{h}\rceil$;

\For{$n = 0,\ldots,N-1$}{
  \For{\text{each} $h_n \in [0, h]$}{
  \vspace{1em}
  \bbh{use the DeepGraphONet's forward pass to compute the prediction
      $$\mathbf{\bar{x}}(t + h_n) = \mathcal{F}_{\theta^*}(\{\mathbf{x}_S(t - \tau_i)\}_{i=1}^m, \{\tau\}_{i=1}^m; S)(h_n).$$}
      \vspace{-1.0em}\;
      append $\mathbf{\bar{x}}(t + h_n)$ to $C$;
      \vspace{1em}
     
  }
 
  \textbf{end for}
  
   update the current time
  $$t \leftarrow t + h;$$
  new observations come in
  $$\{\mathbf{x}_S(t - \tau_i)\}_{i=1}^m;$$

}
  \textbf{end for} 
  
  concatenate $C$;
  
 Return the concatenated predicted trajectories;\;
 \caption{Complete Trajectory Prediction Scheme}
 \label{alg:inference}
\end{algorithm}



\subsection{Experiment 1: Power transient stability} \label{sec:power_exp}
\begin{figure}
    \centering
    \includegraphics[width=\linewidth]{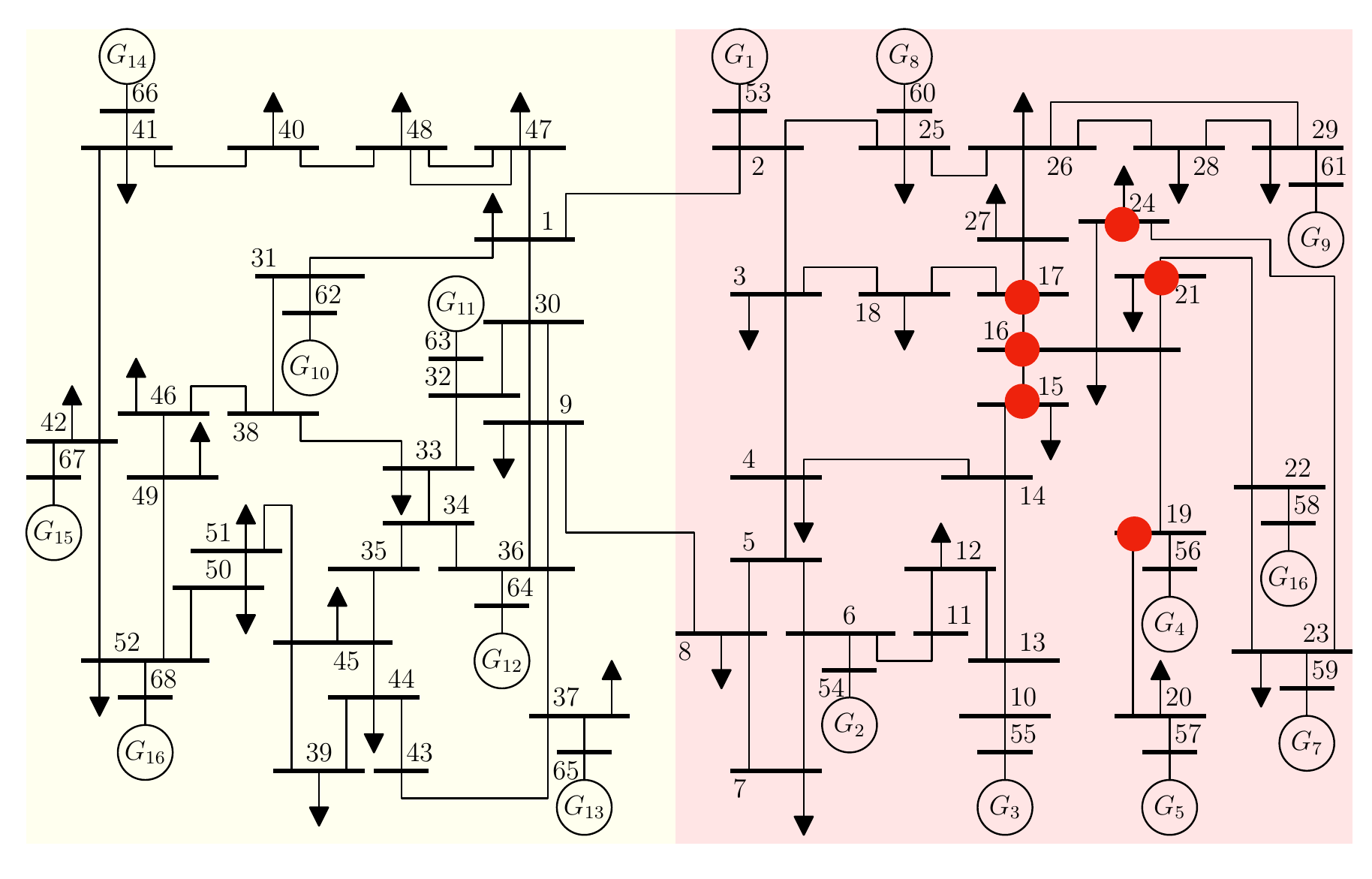}
    \caption{One-line diagram of the New York-New England 16-generator 68-bus power grid. We train our DeepGraphONet using the \textit{arbitrary} subgraph of buses~$S:=\{15,16,17,19,21,24\} \subset G$, highlighted with red dots.}
    \label{fig:powerSys}
\end{figure}
\begin{figure}[t]
\centering
\includegraphics[width=\linewidth]{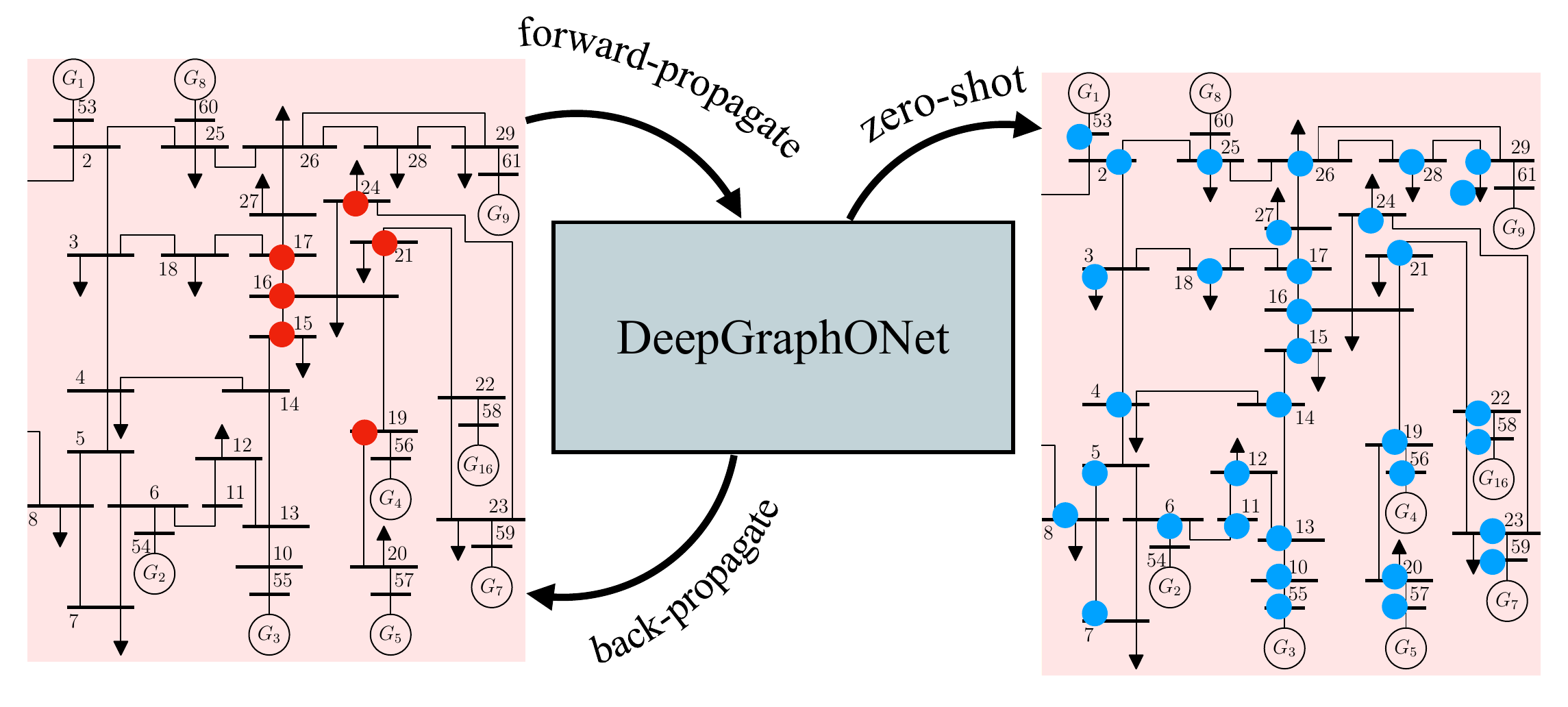}
\caption{Zero-shot learning on a larger subgraph. We train the DeepGraphONet on an arbitrary small subgraph (marked in red) and directly apply the trained network to a much larger subgraph (marked in blue) to achieve high predicting performance~(0.686\% $L_1$ relative error).}
\label{fig:68-bus-system}
\end{figure}

\textit{Dataset.} The power grid dataset consisted of 1830 independent trajectories on a sub-graph with 6 nodes~(red buses in Fig. \ref{fig:68-bus-system}). For each trajectory, there were $701$ time-steps with a resolution of 1~$ms$. For more details about the data generation process, one can refer to~\cite{moya2022deeponet}. Among the available trajectories, $60\%$, $20\%$, and $20\%$ of them were used for training, validation, and testing, respectively. We then applied the trained model to the entire region~(marked in blue in Figure \ref{fig:68-bus-system}), achieving zero-shot learning. 

\begin{figure*}[t]
    \centering
    \includegraphics[width=\linewidth]{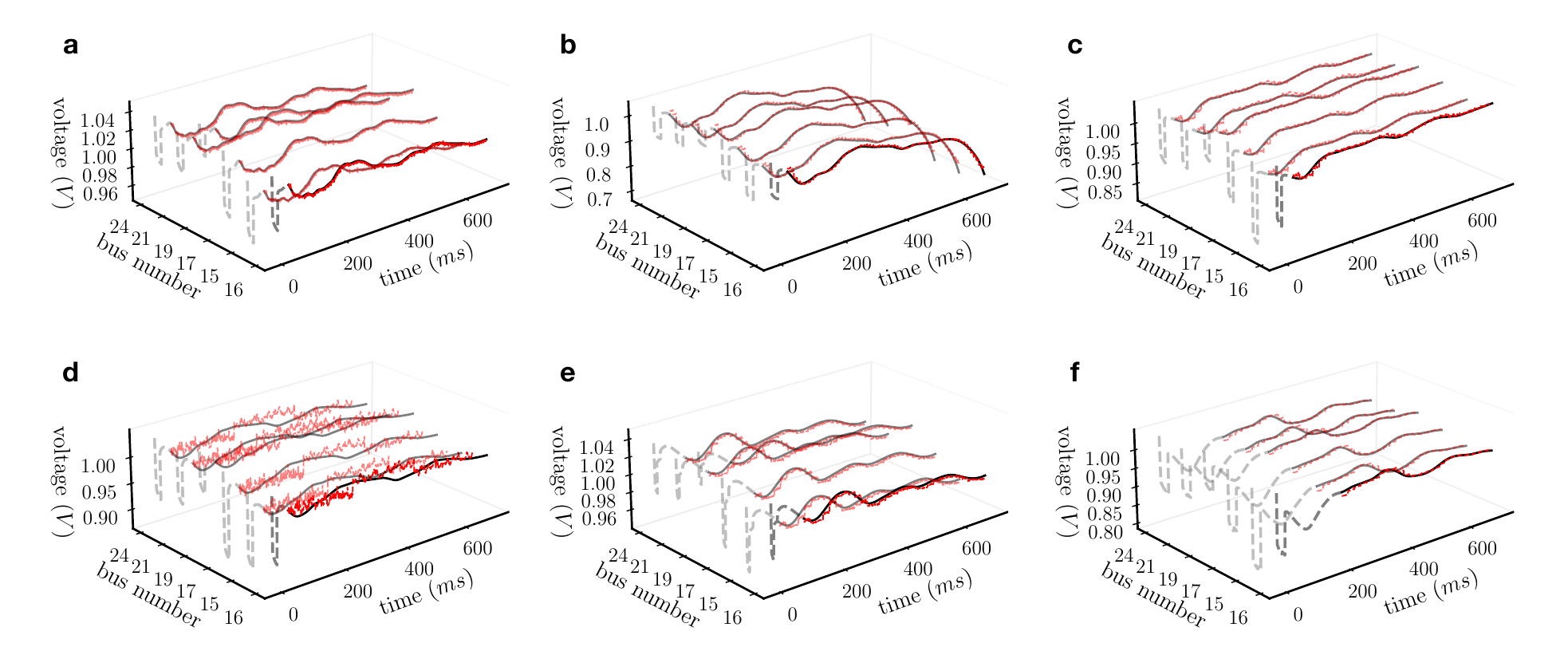}
    \caption{Comparison of the DeepGraphONet predicted transient trajectories (dashed red lines) with the actual trajectories (solid black lines) on the subgraph of buses~$S:=\{15,16,17,19,21,24\} \subset G$. DeepGraphONet generated all predictions at the same time given the memory trajectory (dashed gray lines). We test multiple prediction horizons~$h$ and memory resolutions~$t_M$. In particular, \textbf{a.} $h=2~ms$ with $t_M=50~ms$; \textbf{b.} $h=10~ms$ with $t_M=50~ms$; \textbf{c.} $h=20~ms$ with $t_M=50~ms$; \textbf{d.} $h=200~ms$ with $t_M=50~ms$; \textbf{e.} $h=20~ms$ with $t_M=100~ms$; and \textbf{f.} $h=20~ms$ with $t_M=200~ms$.}
    \label{res1}
\end{figure*}

\begin{figure}[t]
    \centering
    \includegraphics[width=.75\linewidth]{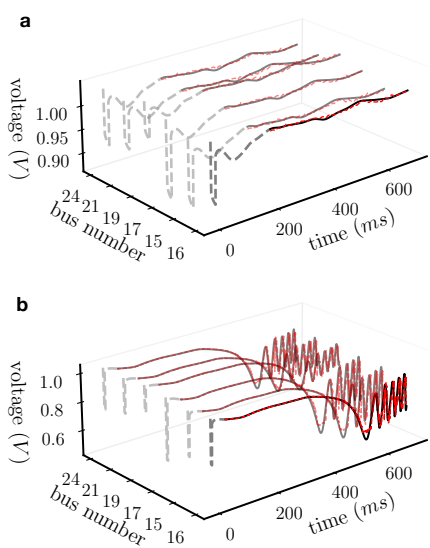}
    \caption{Comparison of the DeepGraphONet predicted transient trajectories (dashed red lines) with the actual trajectories (solid black lines) on the subgraph of buses~$S:=\{15,16,17,19,21,24\} \subset G$ using the proposed resolution-independent DeepGraphONet. We test two time horizons~$h$ and memory resolutions~$t_M$. \textbf{a.}  $h=500~ms$ with $t_M=200~ms$ and \textbf{b.} $h=2~ms$ with $t_M=50~ms$.}
    \label{fig:res-long}
\end{figure}

\textit{Testing Accuracy.} We show the model testing results in Fig. \ref{res1}, Table \ref{t1} and \ref{t2}. Overall, our trained both standard and resolution-independent DeepGraphONets accurately approximated the solution operator for the power grid problem.  The $L_1$ relative errors of all short- and medium-term forecasting settings were within 1\%. Moreover, the resolution-independent DeepGraphONet, taking the input function representation with arbitrary resolution, did not show performance degradation compared to the standard~(Figure \ref{fig:res-long}\textbf{b}). 

\textit{Varying Memory Lengths.} We explored the impact and sufficiency of local memory lengths. We trained the DeepGraphONet with different memory lengths, $t_M=50~ms, t_M=100~ms, t_M=200~ms$, and evaluated the networks on the testing dataset. The prediction results suggest that with larger $t_M$, the average model performance improves~(Fig. \ref{res1}(d)(e)(f)). We observed the same trend in both original and resolution-independent DeepGraphONets, as shown in Table \ref{t2}. 

\textit{Varying Horizons} To investigate the performance change of DeepGraphONet with respect to the change of the output function domain, we varied the prediction horizons $h$. In particular, we trained the networks with $h=2~ms, h=10~ms, h=20~ms, h=200~ms$ with the same memory length, $t_M=50~ms$. The results~(Fig. \ref{res1}\textbf{abc}, and Table \ref{t1}) show for all prediction horizons, both the standard and resolution-independent DeepGraphONet made accurate approximations to the solution operators. Furthermore, the network performance increases as $h$ decreases. For short- and medium-term forecasting, the networks led to $L_1$ relative errors of less than 1\%, whereas when $h=200~ms$, the relative error was over 2\%. However, with longer local memory~($t_M=200~ms$), the trained DeepGraphONet accurately predicted the longer horizon~($h=500~ms$), shown in Fig. \ref{fig:res-long}\textbf{a}.

\begin{table}[]
    \centering
    \caption{The mean and standard deviation~(st.dev.) of the $L_1$-relative error (\%) between the predicted and actual transient response trajectories for multiple time horizons~$h$ and using the (i) \textit{standard} DeepGraphONet and (ii) resolution-independent DeepGraphONet.}
    \begin{tabular}{llcccc}
    \toprule
         $h~(ms)$& & $2$ & $10$ & $20$ & $200$\\
    \midrule
        \multirow{2}{*}{standard}& mean& 0.21 & 0.64 & 0.71 & 2.57\\
         &st.dev.& 0.74 & 2.71 & 2.62 & 4.89 \\ 
    \midrule
        \multirow{2}{*}{res.ind.}&mean& 0.37 & 0.62 & 0.72 & 1.71\\
        &st.dev. & 1.62 &2.57 & 2.61 & 3.58\\
    \bottomrule
    \end{tabular}
    \label{t1}
\end{table}

\begin{table}[]
    \centering
    \caption{The mean and standard deviation~(st.dev.) of the $L_1$-relative error (\%) between the predicted and actual transient response trajectories for multiple memory resolutions~$m$ and using the (i) \textit{standard} DeepGraphONet and (ii) resolution-independent DeepGraphONet.}
    \begin{tabular}{llccc}
    \toprule
          $t_M~(ms)$& & $50$ & $100$ & $200$\\
    \midrule
        \multirow{2}{*}{standard}&mean&  0.71 & 0.70 & 0.66\\
         &st.dev. & 2.62 & 2.82 & 3.14\\
    \midrule
        \multirow{2}{*}{res.ind.}&mean& 0.72 & 0.71 & 0.68\\
        &st.dev.& 2.60 &3.46 & 2.98\\
    \bottomrule
    \end{tabular}
    \label{t2}
\end{table}



\begin{figure}[t]
    \centering
    \includegraphics[width=.75\linewidth, height=4.5cm]{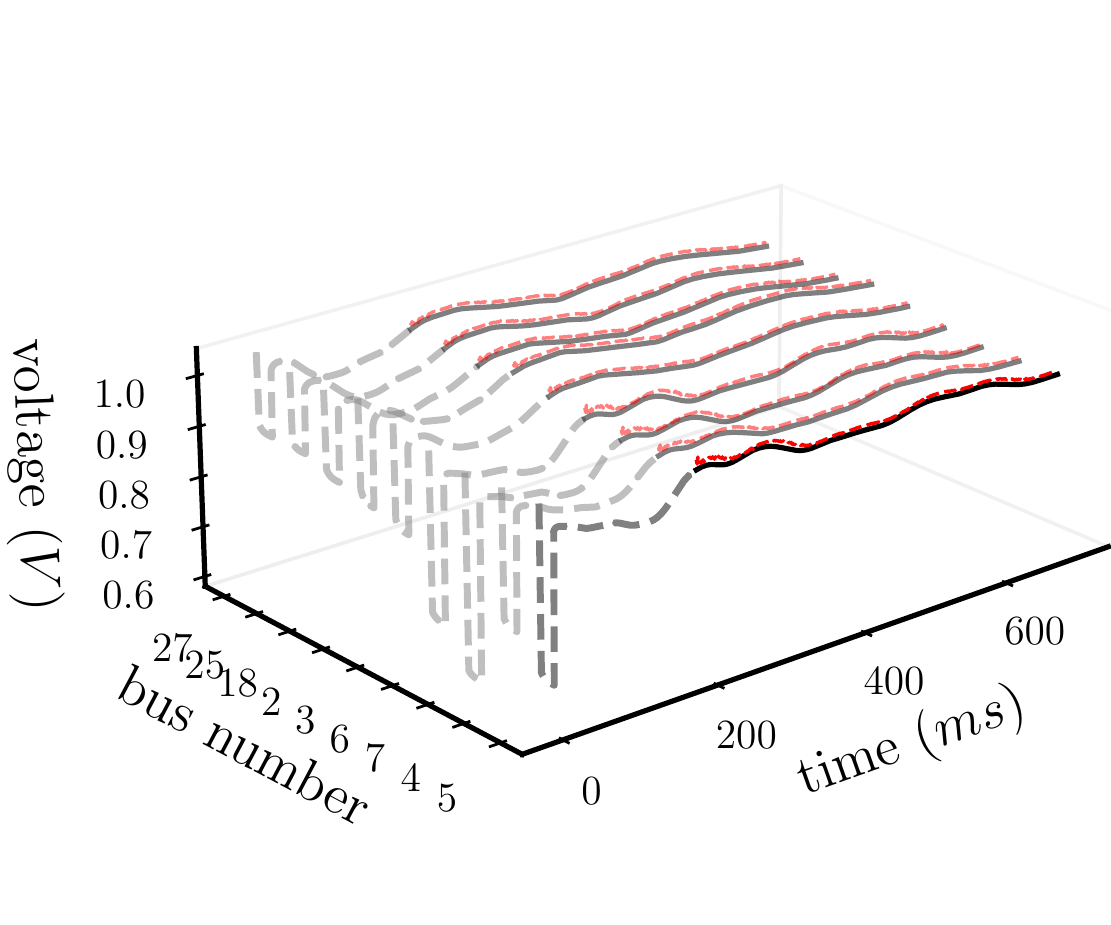}
    \caption{\textit{Zero-shot learning results.} Comparison of the DeepGraphONet predicted transient trajectories (dashed red lines) with the actual trajectories (solid black lines) on a sub-graph of buses~$S':=\{2,3,4,5,6,7,18,25,27\} \subset G$ different from the \textit{training} subgraph~$S$.}
    \label{fig:zero-shot}
\end{figure}

\textit{Zero-shot Learning}
Finally, we applied the trained model to a larger sub-graph from the same power grid model. The new sub-graph has 34 nodes and contains the graph used during training. By directly applying the trained model to the larger sub-graph without further training, we achieved zero-shot learning with high accuracy. Fig. \ref{fig:zero-shot} depicts the prediction results on 9 randomly selected nodes of the new sub-graph. For these nodes, the predictions aligned with the ground truth well with a $L_1$ relative error of 0.686\%. 

With the above, we conclude the proposed DeepGraphONet can successfully learn the solution operator with local memories for predicting the transient stability of power grids. Particularly, the resolution-independent DeepGraphONet, although having the flexibility of taking the discrete input function representation with arbitrary resolution, does not show performance degradation. In addition, one can directly apply the trained DeepGraphONet to solve the power grid transient stability problem with a different underlying graph structure, achieving zero-shot learning.


\subsection{Experiment 2: Traffic forecasting} \label{sec:traffic_exp}
\begin{figure}[t]
    \centering
    \includegraphics[width=.7\linewidth]{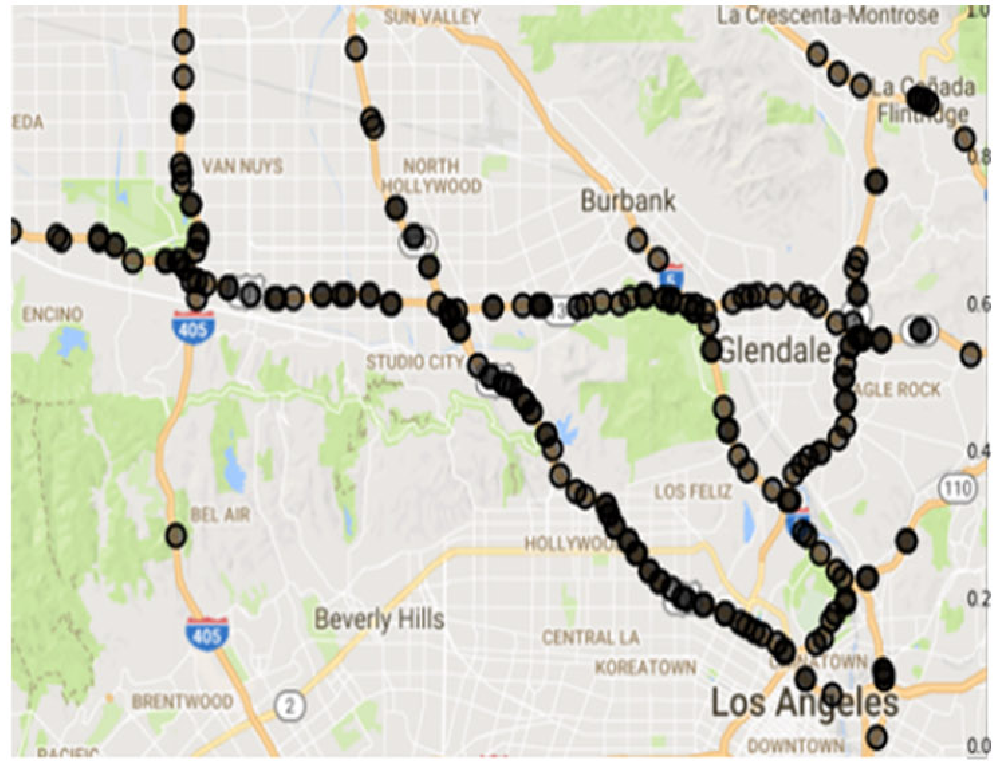}
    \caption{METR-LA traffic network~(adopted from~\cite{li2017diffusion}). The black dots represent the loop detectors, which we regard as the nodes of the graph~$G$.}
    \label{fig:metr-la}
\end{figure}

\textit{Dataset}. We used the METR-LA dataset~\cite{jagadish2014big}, containing 179 loop detectors, shown in Fig. \ref{fig:metr-la}. The dataset contained speed measurements of the year 2018, with a resolution of 5 min. Due to the lack of independent trajectories, we split the data based on the time for training, validation, and testing purposes. We used the data from January to June 2018 for training, July to September 2018 for validation and October to December 2018 for testing. We trained and evaluated the network using the first 20 nodes in the graph and directly applied the trained network to the entire graph for zero-shot learning.

\begin{figure}[t]
    \centering
    \includegraphics[width=1.0\linewidth, height=6.25cm]{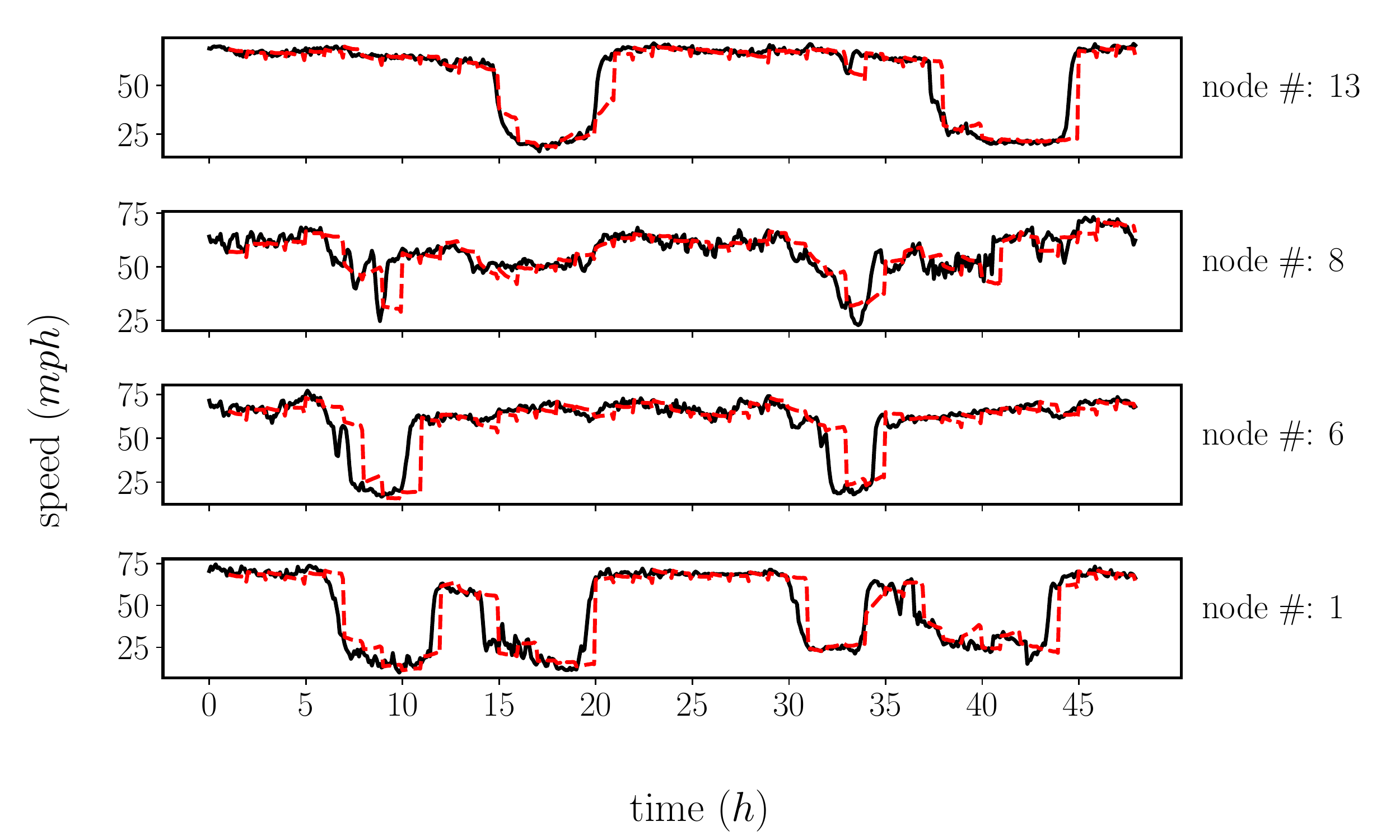}
    \caption{Comparison of the DeepGraphONet predicted trajectories (dashed red lines) with the actual trajectories (solid black lines) on the subgraph~$S:=\{ 1,6,8,13\}$ of the highway network~$G$ using $t_M = 60$ (minutes) memory length.}
    \label{traffic_res}
\end{figure}

\textit{Varying Horizons}. We investigated the DeepGraphONet's performance in traffic forecasting under the commonly used horizons~\cite{wu2020connecting}, $h=15~min, h=30~min, h=60~min$ with the local memory length fixed as $t_M=60~min$. 
Fig. \ref{traffic_res} shows the prediction results from the randomly picked 4 nodes in the graph where the predicting horizon was $h=60~min$. DeepGraphONet captured the future speed of all nodes at the same time accurately using the history observation, while at some sharp turning points, the model did not match the oscillation exactly but captured the average behavior. 
The results suggest that as increasing horizon, the model performance decreased overall and the error range became larger. Table \ref{tab:traffic} lists the mean and standard deviation of the $L_1$ relative error of the models, both for original and resolution-independent DeepGraphONet, with different predicting horizons on the testing set. The results imply the great ability of the proposed DeepGraphONet in predicting the traffic dynamics for all nodes present in the network simultaneously, with $\sim 6$\% error for short-term~(5 min) prediction and $\sim 11$\% for long-term~(1 hour).

\begin{table}
    \centering
     \caption{The mean and standard deviation~(st.dev.) of the $L_1$-relative error (\%) between the predicted and actual traffic trajectories for multiple time horizons~$T$ and using the (i) \textit{standard} DeepGraphONet and (ii) \textit{resolution-independent} DeepGraphONet.}
    \begin{tabular}{llccc}
    \toprule
        $h~(min)$ & & $15$  & $30$ & $60$\\
    \midrule
        \multirow{2}{*}{standard}&mean&  5.52 & 7.60 & 10.79\\
         &st.dev. & 9.49 & 14.74 & 22.05\\
    \midrule
        \multirow{2}{*}{res.ind.}&mean& 6.84 & 8.62 & 11.87\\
        &st.dev.& 12.56 & 16.79 & 23.57\\
    \bottomrule
    \end{tabular}
    \label{tab:traffic}
\end{table}

\begin{figure}[t]
    \centering
    \includegraphics[width=1.0\linewidth, height=6.25cm]{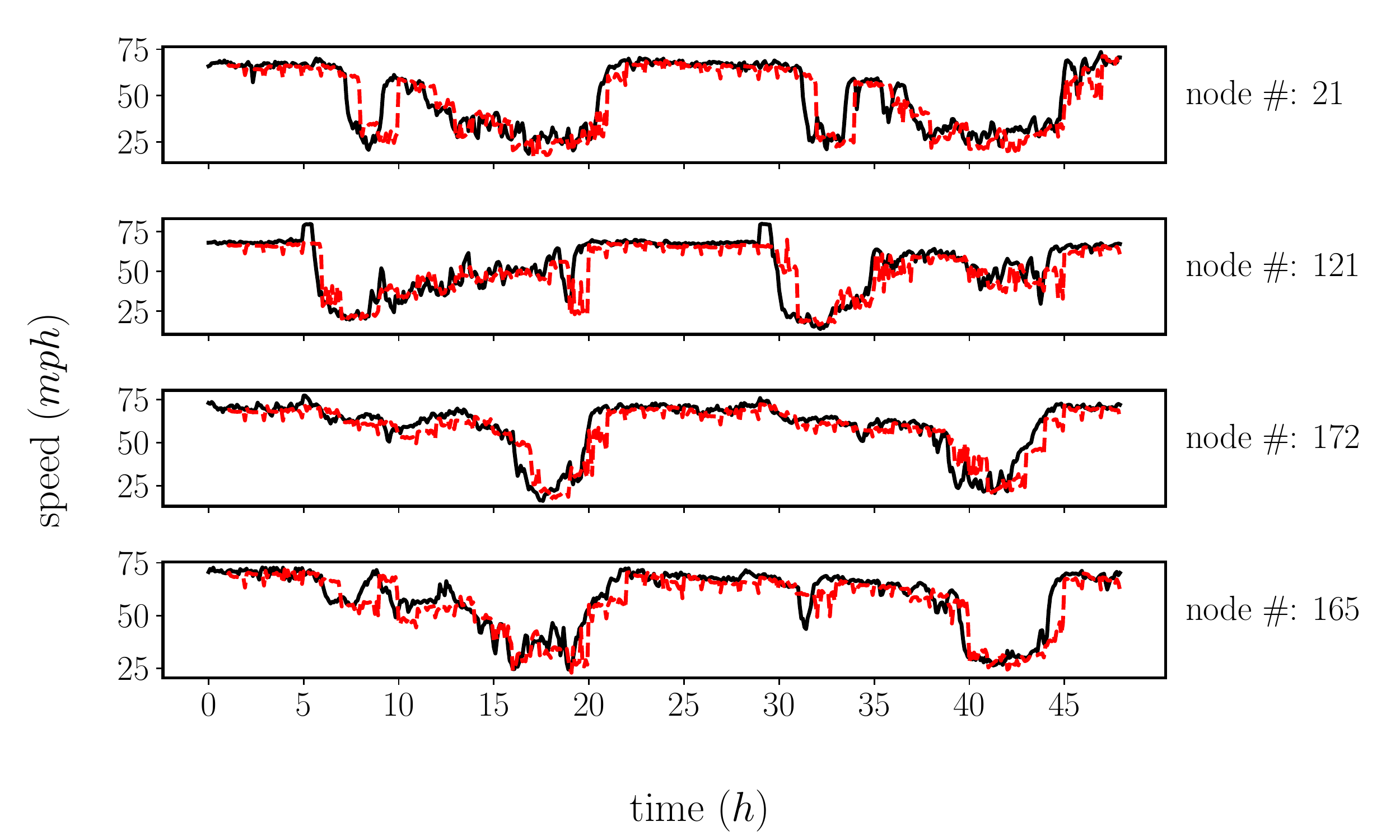}
    \caption{\textit{Zero-shot learning results.} Comparison of the DeepGraphONet predicted trajectories (dashed red lines) with the actual trajectories (solid black lines) on all the 179 nodes of the METR-LA highway network~$G$. We illustrate the predicted trajectories for the following \textit{four} nodes $\{21,121,165,172\}$, which we selected uniformly at random.}
    \label{fig:zero-shot-traffic}
\end{figure}

\paragraph{Zero-shot Learning} To test the DeepGraphONet's potential of zero-shot learning for traffic forecasting, we directly applied the trained network to the entire METR-LA network with 179 nodes. Fig. \ref{fig:zero-shot-traffic} shows the visualized model prediction results on 4 randomly picked nodes among the 179. With 1 hour predicting horizon, the trained model closely predicted the future on the unseen graph structure for all nodes at the same time, resulting in the $L_1$ relative error of  10.79\%. 

The experiment results for the traffic forecasting problem strongly suggest our proposed DeepGraphONet was powerful in learning the solution operator for the traffic dynamics. Also, the DeepGraphONet can be better implemented for practical applications for it is easy to train, can produce predictions at arbitrary time-step within the horizon, and does not require forward dependency to obtain the value at this step.


\section{Discussion and Future Work} \label{sec:discussion}

We used the proposed DeepGraphONet to learn the solution operator of networked dynamical systems and demonstrated its effectiveness in power grid transient stability and traffic forecasting tasks. In particular, the resolution-independent DeepGraphONet demonstrated excellent accuracy in approximating the solution operator without performance degradation while having the ability of flexible discrete input function representation.

We demonstrated that the trained DeepGraphONet can produce predictions at arbitrary query locations within the output function domain and does not require forward dependency in the output sequence. In addition, we showed that one could directly apply the DeepGraphONet, trained on a small sub-graph region, to a larger graph region for the same task, achieving zero-shot learning with high accuracy. With the promising wide-ranged applications of the proposed DeepGraphONet, we provide next a preamble for our future work:
\begin{itemize}
    \item \textit{On Physics-Informed DeepGraphONet.} Our current work focuses on developing a physics-informed version of DeepGraphONet. We plan to use first-principles models to (i) train the DeepGraphONet without data or (ii) use them as prior information during data-driven training.
    \item \textit{On Anomaly Detection with DeepGraphONet.} We will focus part of our future work on using DeepGraphoNet for detecting anomalies. To this end, we must endow the proposed DeepGraphONet with the ability to estimate the network's edges status using graph trajectory data.
    \item \textit{On Learning Networked Control Systems.} We plan to extend the proposed DeepGraphONet for learning networked control dynamical systems. This problem is more challenging because we need to learn the system's response using graph trajectory and its response to external inputs. If successful, our DeepGraphONet for control may become an essential tool for implementing model-based multi-agent continuous reinforcement learning.
    \item \textit{On Predicting Long-term Trajectories.} We plan to develop effective training strategies to alleviate error accumulation such that we can recursively use the trained DeepGraphONet to forecast trajectories for long-term horizons with high accuracy.
\end{itemize}



\label{sec:dis}

\section{Conclusion} \label{sec:conclusion}

In this paper, we introduced the Deep Graph Operator Network~(DeepGraphONet) framework to predict the dynamics of complex systems with underlying sub-graph structures. By fusing the ability from (1) Graph Neural Networks to correlate graph trajectory information and (2) DeepONet to approximate nonlinear operators, we achieved significant results on complex problems such as predicting the (i) transient stability of a control area of a power grid and (ii) the traffic flow within a city or district. Moreover, we built our DeepGraphONet to be resolution-independent, \ie we do not require a set of fixed sensors to encode the graph trajectory history. Finally, we designed a zero-shot learning strategy that enables using the proposed DeepGraphONet on a different sub-graph with high predicting performance.

\section*{Acknowledgment}
This work was supported by the Advanced Grid Modeling Program, Office of Electricity Delivery and Energy Reliability of the U.S. Department of Energy. GL, CM, and YS gratefully acknowledge the support of the National Science Foundation (DMS-1555072, DMS-2053746, and DMS-2134209), Brookhaven National Laboratory Subcontract 382247, and U.S. Department of Energy (DOE) Office of Science Advanced Scientific Computing Research program DE-SC0021142).
\bibliographystyle{ieeetr}
\bibliography{0-main}



\end{document}